%% file: main.tex
\documentclass[11pt]{article}

\usepackage[final]{latex/acl}

\usepackage{times}
\usepackage{latexsym}
\usepackage{enumitem}

\usepackage[T1]{fontenc}
\usepackage[utf8]{inputenc}

\usepackage{microtype}

\usepackage{inconsolata}

\usepackage{graphicx}

\usepackage[frozencache,cachedir=.]{minted}
\usepackage{xcolor}
\usepackage{latex/colab}
\usepackage{latex/symbols}
\usepackage{float}
\usepackage{cleveref}
\usepackage{dirtree}
\usepackage{tcolorbox}
\usepackage{multirow}
\usepackage{booktabs}
\usepackage{pdflscape}

\definecolor{codebg}{HTML}{D9D9D9}
\definecolor{github-dark-bg}{HTML}{0d1117}
\BeforeBeginEnvironment{minted}{\vspace{-3pt}}
\AfterEndEnvironment{minted}{\vspace{-3pt}}
\setminted[python]{
  style=friendly,
  bgcolor=codebg,
  fontsize=\scriptsize,
  autogobble,
  breaklines=true,
breakanywhere=true,
breaksymbolleft=,
breaksymbolright=,
breaksymbolsepleft=0pt,
breaksymbolsepright=0pt,
}
\setminted[bash]{
  style=friendly,
  bgcolor=codebg,
  fontsize=\scriptsize,
  baselinestretch=0.9,
  autogobble,
  breaklines=true,
breakanywhere=true,
breaksymbolleft=,
breaksymbolright=,
breaksymbolsepleft=0pt,
breaksymbolsepright=0pt,
}
\setminted[yaml]{
  style=friendly,
  bgcolor=codebg,
  fontsize=\scriptsize,
  autogobble,
  breaklines=true,
breakanywhere=true,
breaksymbolleft=,
breaksymbolright=,
breaksymbolsepleft=0pt,
breaksymbolsepright=0pt,
}

\newmintinline[pythonline]{python}{fontsize=\footnotesize, bgcolor=codebg,breakanywhere=false}
\newmintinline[bashline]{bash}{fontsize=\footnotesize, bgcolor=codebg,breakanywhere=false,autogobble=false}
\definecolor{cics_red}{HTML}{990000}
\definecolor{cics_light_red}{HTML}{FF7070}
\definecolor{cics_blue}{HTML}{264690}
\definecolor{cics_light_blue}{HTML}{97AEE4}
\definecolor{cics_orange}{HTML}{E94A00}
\definecolor{cics_yellow}{HTML}{EB8F00}
\definecolor{cics_gray}{HTML}{666666}

\title{\xlm{}: A Python package for non-autoregressive language models}

\author{
  \textbf{Dhruvesh Patel}~~
  \textbf{Durga Prasad Maram}\thanks{Equal contribution}~~
  \textbf{Sai Sreenivas Chintha\footnotemark[\value{footnote}]}~~
  \textbf{Benjamin Rozonoyer}~~
\\
  \textbf{Andrew McCallum}
  \\
  University of Massachusetts Amherst
  \\
  \small \{dhruveshpate, dmaram, saisreenivas, brozonoyer, mccallum\}@umass.edu
}

\begin{document}
\maketitle
\input{drafts/eacl_demo/sections/00_abstract}
\input{drafts/eacl_demo/sections/01_introduction}

\input{drafts/eacl_demo/sections/02_related_work}
\input{drafts/eacl_demo/sections/03_xlm}
\input{drafts/eacl_demo/sections/04_demo}
\input{drafts/eacl_demo/sections/07_benchmarks}

\input{drafts/eacl_demo/sections/50_limitations}

\bibliographystyle{latex/acl_natbib}
\bibliography{references}

\input{drafts/eacl_demo/sections/z0_appendix}

\end{document}

%% file: drafts/eacl_demo/sections/00_abstract.tex
\begin{abstract}
In recent years, there has been a resurgence of interest in non-autoregressive text generation in the context of general language modeling. 
Unlike the well-established autoregressive language modeling paradigm, which has a plethora of standard training and inference libraries, implementations of non-autoregressive language modeling have largely been bespoke making it difficult to perform systematic comparisons of different methods.
Moreover, each non-autoregressive language model typically requires it own data collation, loss, and prediction logic, making it challenging to reuse common components.
In this work, we present the \xlm{} python package, which is designed to make implementing small non-autoregressive language models faster.
With a secondary goal of providing a suite of small pre-trained models (through a companion \texttt{xlm-models} package) that can be used by the research community.
\footnote{
\textbf{Python package:} \href{https://pypi.org/project/xlm-core/}{https://pypi.org/project/xlm-core/} and \href{https://pypi.org/project/xlm-models/}{https://pypi.org/project/xlm-models/}. \\
\textbf{Video:} \href{https://drive.google.com/file/d/1uP49n-Gi5sj-Qd6-N1hXaLolEYYKU275/view?usp=drive_link}{https://drive.google.com/file/d/1uP49n-Gi5sj-Qd6-N1hXaLolEYYKU275/view?usp=drive\_link}. \\
\textbf{Source code:}  \href{https://github.com/dhruvdcoder/xlm-core}{https://github.com/dhruvdcoder/xlm-core}. 
}
\end{abstract}

%% file: drafts/eacl_demo/sections/01_introduction.tex
\section{Introduction}\label{sec:introduction}

Autoregressive language models (ARLMs), which generate text sequentially from left to right by adding one token at a time, are well established with a plethora of standard training and inference libraries \citep{wolf2020huggingfacestransformersstateoftheartnatural,olmo20242olmo2furious}.
However, recently, there has been a resurgence of research interest in non-autoregressive text generation due to its potential for faster inference speeds and better generation quality for certain tasks.
Unlike left-to-right generation, non-autoregressive text generation can be done in many ways, for example, using masked diffusion language models ~\citep{sahooSimpleEffectiveMasked2024}, Gaussian diffusion language models~\citep{plaid}, insertion language models ~\citep{ilm}, edit-based language models ~\citep{editflow}, etc.
Moreover, each method typically requires its own data collation, loss, and prediction logic, making it challenging to reuse common components across different methods.
The rapidly expanding landscape of these methods has led to many bespoke implementations, making it extremely difficult to compare them systematically.
In this work, we present the \xlm{} python package, which aims to provide a unified framework for developing and comparing small non-autoregressive language models. 
\xlm{} uses Pytorch~\citep{pytorch} as the deep learning framework, Pytorch Lightning~\citep{pytorch_lightning} for training utilities, and Hydra~\citep{Yadan2019Hydra} for configuration management.
\xlm{} is designed to make implementing small non-autoregressive language models faster without sacrificing flexibility.

The rest of the paper is organized as follows. 
\Cref{sec:design-principles} discusses the core design principles of \xlm{}. In \cref{sec:related_work}, we discuss how \xlm{} serves a unique purpose in the landscape of LLM libraries. \Cref{sec:core-components} presents a high-level overview of the three core components of \xlm{}, followed by
\cref{sec:demonstration} that provides a step-by-step demonstration of how one would implement a new language model using \xlm{}.
Finally, \Cref{sec:benchmarks} presents a set of benchmarking results where we implement three models in \xlm{} to reproduce known results. 

\section{Design Principles}\label{sec:design-principles}
\xlm{} follows the principle of maximal independence. 
The core library provides access to a small number of shared components, which are designed to be model independent, and can be used by any kind of language model.
Each model implementation lives in its own folder/package and is completely independent of other models. This allows researchers to keep their model code clean, self-contained, and easy to share. It also allows them to use their model outside the \xlm{} framework without refactoring. Maximal independence is achieved by following design choices.

\paragraph{Composition over inheritance.} The maximal independence is achieved by using composition over inheritance~\citep{design_patterns}, wherein the core components delegate model specific logic to the specific model instance. For example, as shown in \Cref{fig:overview}, the \datamodule{} carries a collection of \datasetmanager{} \emph{instances}, one for each dataset, and the creation of dataloaders is delegated to the respective \datasetmanager{} instance depending on the stage (train, val, test, or predict).
Similarly, the \harness{} carries instances of \model{}, \lossfunction{}, \predictor{}, to which it delegates the model specific logic for forward pass, loss computation, and generation respectively. Moreover, one can swap out one or more of the four components with a different but compatible implementation without having to change all four.

\paragraph{Copy over branching.} This principle goes against the common wisdom of not copying code. However, in case of research codebases, copying reduces code complexity and helps increase the speed of development~\citep{wolf2020huggingfacestransformersstateoftheartnatural}. 
It naturally creates independence.
Moreover, it also allows templatizing the process of creating a brand new model which helps rapid prototyping by humans as well as LLMs~\citep{llm_api_documentation_accuracy}.

\paragraph{Arbitrary code injection.} Python is a highly dynamic language, which allows one to inject arbitrary code at runtime. 
In production and public facing codebases, this creates a security risk, but this flexibility is a boon for research codebases, as it allows rapid prototyping and experimentation.
As we will discuss in \cref{subsec:configuration_management}, Hydra~\citep{Yadan2019Hydra} provides a powerful mechanism to inject arbitrary code at runtime by allowing one to fill a specific \emph{slot} with an instance of any class.

%% file: drafts/eacl_demo/sections/02_related_work.tex
\section{Related Work}
\label{sec:related_work}

Due to the rapid development of LLMs, there are many libraries for training ~\citep{wolf2020huggingfacestransformersstateoftheartnatural,vonwerra2022trl,litgpt-2023,olmo20242olmo2furious} and inference using auto-regressive LLMs~\citep{kwon2023efficient}.
On the other hand, there are only a handful of python libraries that support non-autoregressive sequence modeling like FairSeq~\citep{ott-etal-2019-fairseq}, and AllenNLP~\citep{Gardner2017AllenNLP}. Moreover, even these libraries are no longer actively maintained and do not support non-autoregressive language modeling.
To the best of our knowledge, \xlm{} is the only library that supports fast prototyping of small non-autoregressive language models, and is geared towards providing a suite of reference implementations for up and coming non-autoregressive language modeling methods.

%% file: drafts/eacl_demo/sections/03_xlm.tex
\section{Core Components of \xlm{}}\label{sec:core-components}

\begin{figure*}[!h]
    \centering
    \includegraphics[width=0.9\textwidth]{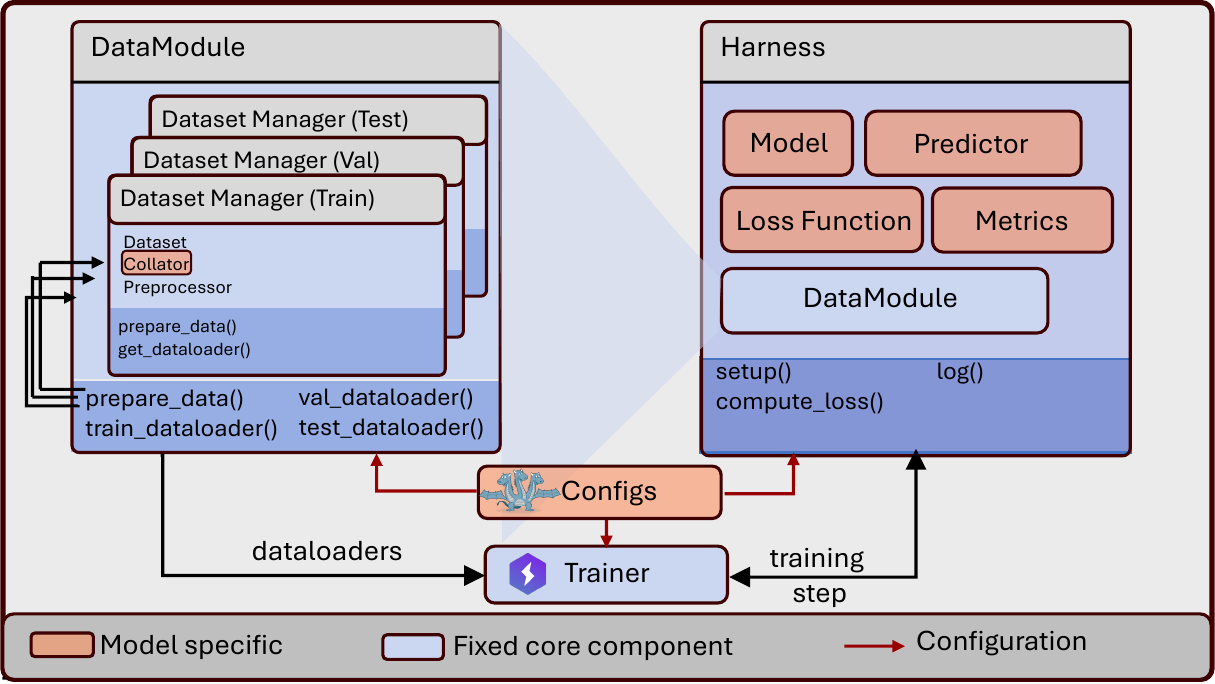}
    \caption{Overview of \xlm{} design. It consists of two classes of components: the core components (\harness{} and \datamodule{}) and the model-specific components, whose implementations depend on the model logic. These components are defined in the configuration files managed by Hydra (see \cref{fig:config}), enabling arbitrary component swapping. The \harness{} component is responsible for instantiating all components (model, loss, predictor, etc.) and delegating their respective functionalities. The \datamodule{} component manages multiple datasets across workflow stages using \datasetmanager{} objects, each handling a dataset and an appropriate \collator{}.}
    \label{fig:overview}
    \vspace{-10pt}
\end{figure*}
In this section, we will discuss the the \harness{}, the  \datamodule{}, and the configuration management, which together handle the execution flow of all the supported {workflows} (\cref{subsec:workflows}) like training, evaluation, prediction and debugging.
In most use cases, these components need not be touched by the user, removing the need for most of the boilerplate code.
\footnote{\xlm{} also has some useful additional features described in \cref{sec:auxiliary_features}.}

\subsection{DataModule}\label{sec:datamodule}
The base \textdatamodule{}, which builds on top of Lightning DataModule, provides a generic, {model and task agnostic interface} for managing arbitrary number of text datasets.
\footnote{See \url{https://lightning.ai/docs/pytorch/stable/data/datamodule.html}. The interface is not specific to text datasets, and can be used for any kind of sequence datasets.}
\durga {DatasetManager per dataset and collator. Maybe we have to reframe.}
This is achieved by using one \datasetmanager{} per dataset as shown in \cref{fig:overview}. 
Each \datasetmanager{} instance is responsible for managing the complete lifecycle of a single dataset, including downloading, preprocessing, caching and managing the data collator and data-loader options. It has slots for the following components that allow injecting custom logic:
\begin{itemize}[leftmargin=*,topsep=0pt,itemsep=0pt,parsep=0pt]
    \item \dataset{} which could point to a HuggingFace dataset or a custom dataset.
    \item \collator{} and \preprocessor{}, both of which can depend on the model type as well as the task.
\end{itemize}

Complete flexibility and independence is achieved by allowing a many-to-many mapping between the \datasetmanager{} instances and the workflow stages (train, val, test, predict), wherein a single \datasetmanager{} instance can be mapped to multiple workflow stages, and vice versa. This ensures that only a single copy of the dataset is loaded into memory but if needed it can be used at multiple places in the workflow.

The core implementation of \datasetmanager{} supports all common training strategies for small models: single-node single-GPU, single-node multi-GPU and multi-node multi-GPU, with map style and iterable style dataset support for each. The user simply needs to provide the respective arguments in the config file.

\subsection{Harness}\label{sec:harness}
The \harness{} is the main class that inherits from the PyTorch lightning's \lightningmodule{}
\footnote{\url{https://lightning.ai/docs/pytorch/stable/common/lightning_module.html}}, and is responsible for instantiating all the components like the model, loss function, predictor, etc., based on the configuration files.  
As shown in \cref{fig:overview}, the \harness{} has slots (attributes) for all the core components, and it delegates the model specific logic to the respective components' methods.
Inheriting from the \lightningmodule{} allows us to use all the features of PyTorch lightning, such as logging, checkpointing, saving, etc., and also allows us to use the \texttt{LightningTrainer}. See \cref{sec:harness_appendix} for more details.

\subsection{Configuration Management}
\label{subsec:configuration_management}

In \xlm{}, the configuration files have two roles. First, like any other configuration system (e.g. Python's \pythonline{ArgParse} 
), it allows the user to specify various parameter values that dictate the behavior of the system. 
Second, through the use of \pythonline{hydra.utils.instantiate}, it allows swapping out entire components directly from the configuration file, without changing a single line of python code. This enables arbitrary code injection at runtime.
Hydra configs themselves can be arbitrarily nested, and one config file can be referred in another config file, the composition of which is automatically taken care of by Hydra. 
This enables a modularization of the configuration files themselves.\footnote{Please refer to the Hydra documentation for more details \url{https://hydra.cc/docs/intro/}.}

%% file: drafts/eacl_demo/sections/04_demo.tex
\section{Demonstration}
\label{sec:demonstration}
In this section, we will walk through, step by step, the process of implementing a new language model using the \xlm{} library. 
In order to demonstrate the flexibility of the library, we pick a non-standard language modeling paradigm for this demonstration. Specifically, we will implement the Insertion Language Model (ILM) of \citet{ilm}, which generates text by iteratively inserting tokens in the existing sequence by selecting the position and the vocabulary item to insert.
In order to keep the demonstration simple, we will use the synthetic seq2seq task of generating a path on a star shaped graph as the task~\citep{ilm} on the \texttt{StarEasy} dataset \textsuperscript{\ref{tab:resources}} .
We also provide a demonstration of training ILM on the LM1B corpus ~\citep{chelba2014billionwordbenchmarkmeasuring} in Appendix \ref{sec:demonstration_on_lm1b}.

\begin{figure}[t]
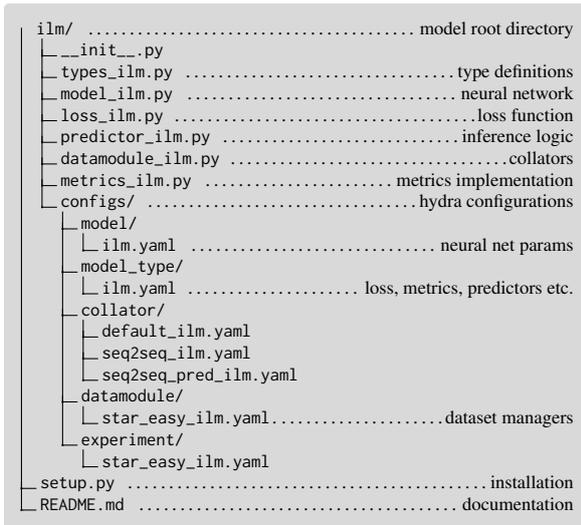

\begin{tcolorbox}[colback=codebg, colframe=codebg, boxrule=0pt, arc=2pt, left=3pt, right=3pt, top=3pt, bottom=3pt, fontupper=\scriptsize]
\scriptsize
\DTsetlength{0.1em}{0.8em}{0.2em}{0.4pt}{0.1em}
\dirtree{%
.1 ilm/ \DTcomment{model root directory}.
.2 \_\_init\_\_.py.
.2 types\_ilm.py \DTcomment{type definitions}.
.2 model\_ilm.py \DTcomment{neural network}.
.2 loss\_ilm.py \DTcomment{loss function}.
.2 predictor\_ilm.py \DTcomment{inference logic}.
.2 datamodule\_ilm.py \DTcomment{collators}.
.2 metrics\_ilm.py \DTcomment{metrics implementation}.
.2 configs/ \DTcomment{hydra configurations}.
.3 model/.
.4 ilm.yaml \DTcomment{neural net params}.
.3 model\_type/.
.4 ilm.yaml \DTcomment{ loss, metrics, predictors etc.}.
.3 collator/.
.4 default\_ilm.yaml.
.4 seq2seq\_ilm.yaml.
.4 seq2seq\_pred\_ilm.yaml.
.3 datamodule/.
.4 star\_easy\_ilm.yaml.\DTcomment{dataset managers}.
.3 experiment/.
.4 star\_easy\_ilm.yaml.
.1 setup.py \DTcomment{installation}.
.1 README.md \DTcomment{documentation}.
}
\end{tcolorbox}
\vspace{-10pt}
\caption{Directory structure generated by the scaffolding script.}
\vspace{-10pt}
\label{fig:directory_structure}
\end{figure}

To construct a new model, we need to create a fresh directory for the model, which will contain the implementations of the model specific components: \model{}, \lossfunction{}, \predictor{}, and \collator s. It will also contain their respective configuration files.
In order to make the process of creating a new model easier, we provide a scaffolding script that automatically generates the necessary files. It can be invoked by executing
\bashline{xlm-scaffold ilm}
which will generate the directory structure as shown in \cref{fig:directory_structure}.

The scaffold files already contain placeholder empty class and function declarations. Next we will walk through the process of adding implementation for each of the python files shown in \cref{fig:directory_structure}. Finally, we will show how to reference these components configuration files.

\begin{figure*}[ht!]
    \centering
    \includegraphics[width=\textwidth]
    {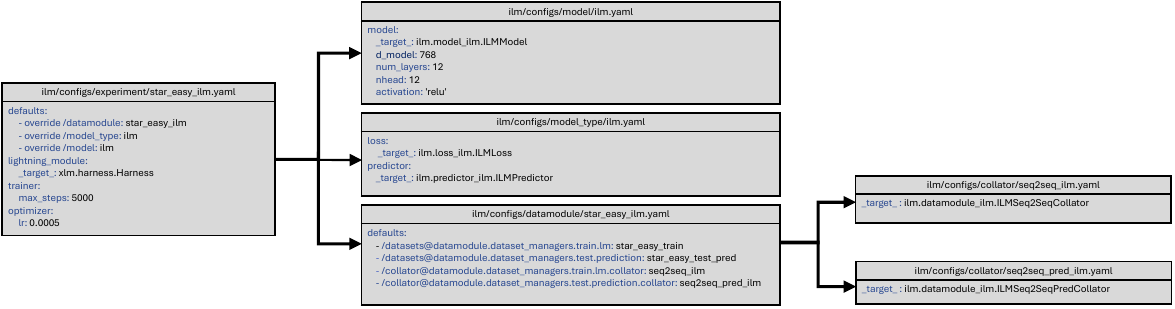}
    \vspace{-15pt}
    \caption{Configuration tree for a typical experiment (e.g. for ILM for a seq2seq planning task on the \texttt{StarEasy} dataset). The \bashline{experiment} config is at the root of the nesting structure, contains global parameters, and composes component configs (\bashline{model}, \bashline{model_type}, and \bashline{datamodule}). The \bashline{model/ilm.yaml} file stores the parameters for the model class. The \bashline{model_type/ilm.yaml} file contains the information needed to instantiate the loss function, predictor, and metric components. The \bashline{datamodule/star_easy_ilm.yaml} composes the configs of the \datasetmanager s and \collator s (here, \texttt{StarEasy} and seq2seq collators).
   Note: Only partial entries are shown in the figure for brevity.  }
    \label{fig:config}
    \vspace{-15pt}
\end{figure*}

\subsection{Model}
In \bashline{model_ilm.py}, we define the neural network backbone for the model which inherits from \pythonline{torch.nn.Module} and implements the \pythonline{forward()} method. We have the complete freedom to decide the arguments that the \pythonline{forward()} method takes.

\begin{minted}{python}
from xlm.model import Model
from xlm.modules.rotary_transformer import 
    RotaryTransformerLayer

class ILMModel(torch.nn.Module, Model):
    ...
    def forward(self, input_ids, attention_mask,...):
        self.encoder_layer = RotaryTransformerLayer(
            d_model,
            nhead,
            dim_feedforward,
            dropout,
            activation,
            layer_norm_eps,
        )
        ...
        return vocab_logits, stopping_logits
\end{minted}
Many of the \xlm{} modules detailed in section \ref{para:modules} can be used for building the architecture like the use of \pythonline{RotaryTransformerLayer} here for the encoder. 
After defining the model class, the constructor arguments needed for default instantiation along with the fully qualified class path are stored in the config \bashline{configs/model/ilm.yaml} file: 
\begin{minted}{yaml}
# @package _global_
model:
  _target_: ilm.model_ilm.ILMModel
  ...
\end{minted}
\vspace{-10pt}
\subsection{LossFunction}
The \lossfunction{} is a callable that takes in a batch of inputs and returns a dictionary of values, with a mandatory "loss" key and any other optional values that we want to log.
In ILM, the loss function computes two components: (1) a cross-entropy loss over only the positions where tokens were dropped, and (2) a binary classification loss that predicts whether input sequence is complete. 
\vspace{-5pt}
\begin{minted}{python}
from xlm.harness import LossFunction
from types_ilm import ILMBatch, ILMLossDict

class ILMLoss(LossFunction[ILMBatch, ILMLossDict]):
    def loss_fn(self, batch: ILMBatch, ...) -> ILMLossDict:
        vocab_logits, stopping_logits = self.model(**batch)
        
        vocab_logit_loss = masked_cross_entropy(vocab_logits, batch["target_ids"])
        
        stopping_loss = binary_cross_entropy(stopping_logits)
        return {"loss": vocab_logit_loss + stopping_loss, ...}    
\end{minted}
The default loss parameters are stored in the config \bashline{configs/model_type/ilm.yaml} file under the \bashline{loss} key as shown in \cref{fig:model_type_config}.

\subsection{Data Pipeline}
To setup the data pipeline, we just need to configure the dataset and implement model specific collators for each dataset, and add the configuration files for the \datasetmanager s and \collator s.
The \xlm{} comes with some synthetic seq2seq, and language modeling datasets preconfigured. This includes the synthetic \texttt{StarEasy} dataset. 
\footnote{See \cref{sec:appendix_new_datasets} for learning how to configure a new dataset, and \cref{sec:preconfigured-tasks} for the list of preconfigured tasks.}
For this demonstration, we will use the preconfigured \texttt{StarEasy} dataset, wherein each example consists of a prompt and a target path. The prompt contains an edge list (in random order) of a star graph and the start and end nodes. The target contains the gold path from the start node to the end node \citep{ilm}. An example prompt and target is shown in \cref{fig:star_easy_example}.
\begin{figure}[t]
\begin{minted}{python}
edge_list = [[1, 5], [1, 7], [7, 9]]
source = 1
goal = 9
path = [1, 7, 9]
sequence = "CLS 1 7 1 5 7 9 1 9 BOS 1 7 9 PAD PAD"
\end{minted}
\vspace{-20pt}
\caption{An example of a prompt and target for the \texttt{StarEasy} dataset.}
\vspace{-10pt}
\label{fig:star_easy_example}
\end{figure}

\subsection{Collators}
A \collator{} is a callable that takes in a list of raw examples from the dataset and returns a batch to be fed to the \lossfunction{} or the \predictor{}.
Typically, a \collator's implementation depends on the model type and the type of task, e.g. the collator for a seq2seq task will be different from that of unconditional language modeling task.
Moreover, a collator for loss computation will be different from that used for prediction. For the synthetic seq2seq task of star graphs, we need to implement two collators: one for training and one for prediction. 
The training collator randomly drops tokens from the gold path and places the dropped tokens under the \pythonline{"target_ids"} key in the batch. The prediction collator only keep the prompt under the \pythonline{"input_ids"} key in the batch leaving all the tokens in the path to be predicted by the model.
\begin{minted}{python}
    from xlm.datamodule import Collator
    from types_ilm import ILMBatch
    
    class ILMSeq2SeqCollator(Collator):
        def __call__(self, examples: List) -> ILMBatch:
            prefix = prepare_prefix_ids([e["prompt_ids"] for e in examples])
            suffix = drop_tokens([e["input_ids"] for e in examples])
            return {
                "input_ids": torch.cat([prefix["input_ids"], suffix["input_ids"]], dim=1),
                "target_ids": suffix["target_ids"],
                "n_drops": suffix["n_drops"]
            }
    
    class ILMSeq2SeqPredCollator(ILMSeq2SeqCollator):
        def __call__(self, examples: List) -> ILMBatch:
            prefix = prepare_prefix_ids([e["prompt_ids"] for e in examples])
            target_ids = prepare_target_ids_for_test([e["input_ids"] for e in examples])
            return {
                "input_ids": prefix["input_ids"],
                "target_ids": target_ids["target_ids"],
                "n_drops": None
            }
    \end{minted}
The collators are configured by adding \bashline{configs/collator/seq2seq_ilm.yaml}: 
\begin{minted}{yaml}
_target_: ilm.datamodule_ilm.ILMSeq2SeqCollator
\end{minted}
and \bashline{configs/collator/seq2seq_pred_ilm.yaml}:
\begin{minted}{yaml}
_target_: ilm.datamodule_ilm.ILMSeq2SeqPredCollator
\end{minted}
Finally, the entire data pipeline is configured by adding \bashline{configs/datamodule/star_easy_ilm.yaml} config file as shown in \cref{fig:datamodule_config}.

\subsection{Predictor}
The \predictor{} is a callable that implements the inference loop for the model.
In ILM, the \predictor{} implements iterative infilling by repeatedly sampling the location of insertion and the vocabulary item to insert till the stopping classification head indicates that generation should stop.

\begin{minted} {python}
from xlm.harness import Predictor
from types_ilm import ILMBatch,ILMPredictionDict

class ILMPredictor(Predictor[ILMBatch, ILMPredictionDict]):
    def predict(self, batch: ILMBatch, ...) -> ILMPredictionDict:
        ...
        return {"text": decoded_texts, "history": generation_history}
\end{minted}
\xlm{} automatically invokes the predictor during evaluation and handles decoding and metric computation. The predictor class path is stored in the config \bashline{configs/model_type/ilm.yaml} file under the \bashline{predictor} key as shown in \cref{fig:model_type_config}
\subsection{Experiment Configuration}
\label{subsec:experiment_config}
Once the individual components are implemented and configured as elaborated previously, the hierarchical configuration tree is formed through the \emph{experiment config}. \Cref{fig:config} depicts the configuration tree for the ILM model on the \texttt{StarEasy} dataset. 
The arrows in the figure indicate the nesting structure: the main experiment config \bashline{experiments/star_easy_ilm.yaml} refers to the \texttt{model} \bashline{model/ilm.yaml}, \texttt{model\_type} \bashline{model_type/ilm.yaml}, and the \texttt{datamodule} \bashline{datamodule/star_easy_ilm.yaml}, which in turn refers to the the collators.

\subsection{Workflows}
\label{subsec:workflows}
After creating the experiment configuration, we are ready to use the model.
\xlm{} provides three main workflows, corresponding to training, evaluation, and inference. 
They can be run using the following command
\begin{minted}{bash}
$ xlm job_type=[JOB_TYPE] job_name=[JOB_NAME] experiment=[CONFIG_PATH]
\end{minted}
The \bashline{job_type} argument can be one of \bashline{train} , \bashline{eval} and \bashline{generate}. 
The name of the experiment config file (without the \bashline{.yaml} extension) containing all necessary overrides is given under the \bashline{experiment} option. 
\xlm{} also provides a group of debug settings that can be used to perform a quick debug run that tries to overfit on a single batch of data.
This can be used by appending the \bashline{debug=overfit} option to the command.
Finally, the model code can be packaged as a standalone python package.
\vspace{-5pt}
\begin{minted}{bash}
# Packaging
python setup.py sdist bdist_wheel
# Installation
pip install ilm
\end{minted}
Post training, the model weights can be extracted and uploaded to the Hugging Face model hub (see \cref{sec:auxiliary_features} for the details).

%% file: drafts/eacl_demo/sections/07_benchmarks.tex
\section{Benchmarks}
\label{sec:benchmarks}
\xlm{} is a framework aimed at making research prototyping of small non-autoregressive language models easier.
The purpose of this section is to show that the library can be used to reproduced known results for small non-autoregressive language models. 
As representative tasks, we pick one seq2seq task and one unconditional language modeling task.
Specifically, we reproduce the results on the synthetic seq2seq of path finding on star graphs and unconditional language modeling tasks on LM1B.

\noindent\textbf{Synthetic Planning Tasks}
We use the hyperparameters reported in for training auto-regressive model (ARLM), masked diffusion model (MDLM)\citep{sahooSimpleEffectiveMasked2024}, masked language model (MLM) and insertion language model (ILM) on the three variants of the synthetic planning tasks \citep{ilm}.  The results (\cref{tab:star_benchmarks}) are within 2\% of the reported results in the original papers.

\begin{table}[t]
    \setlength{\tabcolsep}{3.5pt}
           \centering
           \caption{Benchmark performance on planning seq2seq task on star graphs. The columns represent token and sequence accuracies for each model.
           }
           \label{tab:star_benchmarks}
           \vspace{-5pt}
           \resizebox{\linewidth}{!}{
           \begin{tabular}{@{}lcccccc@{}}
               \toprule
               \multirow{2}{*}{Model}  & \multicolumn{2}{c}{\small Easy} & \multicolumn{2}{c}{\small Medium} & \multicolumn{2}{c}{\small Hard}                                                 \\\cmidrule{2-7}
                                       & {\small Seq.}                                          & {\small Token }                                            & {\small Seq.}                                           & {\small Token }     & {\small Seq.}    & {\small Token }        \\ \midrule
               ARLM                     & 33.1                                                & 81.7                                                  & 77.2                                                 & 82.1           & 25.2          & 43.7                \\
               ILM                     & 100.0                                               & 100.0                                                 & {100.0}                                       & {100.0} & {97.5} & {98.2} \\ 
                MLM                     & 100.0                                               & 100.0                                                 & 83.1                                                 & 98.0           & 25.3          & 79.6                \\
               MDLM                     & 100.0                                               & 100.0                                                 & 36.5                                                 & 90.6           & 21.0          & 54.9                \\
\bottomrule
           \end{tabular}%
           }
           \vspace{-2pt}
    \end{table}

\noindent\textbf{Language Modeling}
We train a 12 layer transformer as ARLM, MDLM, and ILM, respectively, on the  LM1B corpus~\citep{chelba2014billionwordbenchmarkmeasuring}.
In order to make the results comparable, we use negative  loglikelihood under Llama 3.2 8B model as the metric. 
The results are reported in \cref{tab:lm-benchmarks}, which are close to the results reported in \citet{ilm} for the same settings.
\begin{table}[t]
\caption{Benchmark performance on unconditional language modeling on LM1B. The rows represent negative loglikelihood (under Llama 3.2) and entropy of the generated sequences for each model.
    }
    \label{tab:lm-benchmarks}
    \vspace{-5pt}
    \begin{tabular}{lcccc}
        \toprule
         & \small{corpus} & ARLM & MDLM & ILM  \\
        \midrule
        NLL      & \small{3.71} & {3.94} & 4.81 & {4.72} \\
        Entropy      & \small{3.08} & {3.12} & {3.70} & 2.81 \\
        \bottomrule
    \end{tabular}%
    \vspace{-8pt}
\end{table}

%% file: drafts/eacl_demo/sections/50_limitations.tex
\section{Conclusion and Future Work}
\label{sec:conclusion_and_future_work}
We presented a modular python package that makes prototyping small non-autoregressive language models easier. We plan on adding reference implementations and benchmark more models \citep{editflow} as well as add support for non-text sequence generation tasks like molecule generation (see \cref{app:planned features} for details).

%% file: drafts/eacl_demo/sections/z0_appendix.tex
\appendix

\section{Demonstration on LM1B}\label{sec:demonstration_on_lm1b}
In \cref{sec:demonstration}, we demonstrated ILM on a seq2seq task where the model generates a path given a graph description as a prompt. 
Here, we show how to implement ILM for unconditional language modeling on a real-world text corpus (LM1B), where there is no prompt and the model generates text from scratch.
The steps translate to any other type of language model.

\subsection{Collator for Unconditional LM}
The first difference from the seq2seq task is the collator. 
In the case of ILM, instead of \pythonline{ILMSeq2SeqCollator}, which protects the prompt tokens from being dropped, we need to implement a new collator, which we call \pythonline{DefaultILMCollator}, which drops tokens uniformly from the entire sequence:
\begin{minted}{python}
from xlm.datamodule import Collator

class DefaultILMCollator(Collator):
    """Used for pre-training."""
    
    def __call__(self, examples: List) -> ILMBatch:
        batch = ilm_single_segment_collate_target_fn(
            [e["input_ids"] for e in examples],
            ...
            sample_n_drops_fn=_n_drop_uniformly,  # uniform drop count
            drop_indices_fn=_drop_uniformly,       # uniform drop positions
        )
        return batch
\end{minted}

\subsection{Datamodule Configuration}
The datamodule configuration for unconditional language modeling differs from the seq2seq task at two places: (1) we need to swap out the collator configuration to use our new \pythonline{DefaultILMCollator}, and (2) we need to add a new dataset manager for unconditional generation, which manages dataset of blank sequences.
\begin{minted}{yaml}
# configs/datamodule/lm1b_ilm.yaml
# @package _global_
defaults:
  - /datasets@datamodule.dataset_managers.train.lm: lm1b_train
  - /datasets@datamodule.dataset_managers.val.lm: lm1b_test
  - /datasets@datamodule.dataset_managers.val.unconditional_prediction: 
      text_unconditional_prediction
  - /collator@datamodule.dataset_managers.train.lm.collator: default_ilm
  - /collator@datamodule.dataset_managers.val.lm.collator: default_ilm
  - /collator@datamodule.dataset_managers.val.unconditional_prediction.collator: 
      default_ilm
\end{minted}
For unconditional generation, the model must generate text starting from an empty sequence. This is handled by \pythonline{ILMEmptyDataset}, which generates empty examples that the model fills entirely:
\begin{minted}{python}
class ILMEmptyDataset(IterableDataset):
    def __init__(self, tokenizer: Tokenizer, num_examples: int):
        self.tokenizer = tokenizer
        self.num_examples = num_examples

    def __iter__(self):
        for _ in range(self.num_examples):
            yield self.tokenizer("", add_special_tokens=False)
\end{minted}
This dataset is referenced in the experiment configuration:
\begin{minted}{yaml}
# configs/experiment/lm1b_ilm.yaml
datamodule:
  dataset_managers:
    val:
      unconditional_prediction:
        dataset_constructor_str: ilm.datamodule_ilm.ILMEmptyDataset
\end{minted}

\section{Hydra Configs}
\Cref{fig:model_type_config} and \cref{fig:datamodule_config} and  show the model type and datamodule configs, respectively.
\begin{figure}[t]
    \begin{minted}{yaml}
# @package _global_
loss:
  _target_: ilm.loss_ilm.ILMLoss
  ... # other arguments

predictor:
  _target_: ilm.predictor_ilm.ILMPredictor
  ... # other arguments

reported_metrics:
  train: # reported during training loop
    lm: # dataloader name
      accumulated_loss: # metric name
        prefix: train/lm # str prefix for logging
        update_fn: ilm.metrics_ilm.mean_metric_update_fn # callable
    ... # metrics for additional dataloaders
  val: 
    ...
  test:
    ...
    \end{minted}
\caption{The \bashline{configs/model_type/ilm.yaml} config file for the ILM model. It contains sections for \lossfunction{}, \predictor{} and \metrics{}.}
\label{fig:model_type_config}
\end{figure}
\begin{figure}[t]
\begin{minted}{yaml}
# @package _global_
defaults:
  - /datasets@datamodule.dataset_managers.train.lm: star_easy_train
  - /datasets@datamodule.dataset_managers.val.lm: star_easy_val
  - /datasets@datamodule.dataset_managers.val.prediction: star_easy_val_pred
  - /datasets@datamodule.dataset_managers.test.lm: star_easy_test
  - /datasets@datamodule.dataset_managers.test.prediction: star_easy_test_pred
  - /datasets@datamodule.dataset_managers.predict.prediction: star_easy_test_pred
  - /collator@datamodule.dataset_managers.train.lm.collator: seq2seq_ilm
  - /collator@datamodule.dataset_managers.val.lm.collator: seq2seq_ilm
  - /collator@datamodule.dataset_managers.val.prediction.collator: seq2seq_pred_ilm
  - /collator@datamodule.dataset_managers.test.lm.collator: seq2seq_ilm
  - /collator@datamodule.dataset_managers.test.prediction.collator: seq2seq_pred_ilm
  - /collator@datamodule.dataset_managers.predict.prediction.collator: seq2seq_pred_ilm
...
\end{minted}
\caption{The \bashline{configs/datamodule/star_easy_ilm.yaml} config file for the ILM model. It contains sections for datasetmanagers and collators.}
\label{fig:datamodule_config}
\end{figure}
\durga {We might wanna explain lm and prediction keys.}
\section{Harness}
\label{sec:harness_appendix}
This section describes the functionality implemented in \harness{}.
\subsection{Loggers}
Logger components can be registered through the \bashline{loggers} key. \xlm{} provides pre-configured tensorBoard and WandB logger configurations, with tensorboard being the default. However, all PyTorch Lightning–supported loggers can also be used.
\begin{figure}[H]
\begin{minted}{yaml}
defaults:
  - override /loggers: 
    -wandb
\end{minted}
\caption{The entries for loggers in the \bashline{experiment} config file.}
\label{fig:loggers}
\end{figure}

\subsection{Logging Metrics}
The library provides various preconfigured metrics for different stages, such as \textbf{accumulated loss} (mean loss value), \textbf{exact match}, and \textbf{token accuracy}. Each of these metric components inherits from the \bashline{torchmetrics.Metric } class and is wrapped by default using the \bashline{xlm.metrics.MetricWrapper} module, which manages the computation of its value. Another key method to define is the \bashline{update_fn} function, which takes raw input batch sequences and loss function outputs, and transforms them into a dictionary of entries used by the Metric class to compute the final value. This allows for customization, enabling custom metric logic depending on the model and task. Different metrics can be configured for various workflow stages as depicted in \cref{fig:metrics}.
\begin{figure}[t]
    \begin{minted}{yaml}
reported_metrics:
  train:
    lm:
      accumulated_loss:
        _target_: xlm.metrics.MetricWrapper
        name: accumulated_loss
        metric:
          _target_: torchmetrics.MeanMetric
        prefix: train/lm
        update_fn: xlm.lm.ilm.metrics_ilm.mean_metric_update_fn
  val:
    ...
  test:
    lm:
      accumulated_loss:
        _target_: xlm.metrics.MetricWrapper
        name: accumulated_loss
        metric:
          _target_: torchmetrics.MeanMetric
        prefix: test/lm
        update_fn: xlm.lm.ilm.metrics_ilm.mean_metric_update_fn
    prediction:
      exact_match:
        _target_: xlm.metrics.MetricWrapper
        name: exact_match
        metric:
          _target_: xlm.metrics.ExactMatch
        prefix: test/prediction
        update_fn: xlm.metrics.seq2seq_exact_match_update_fn
      token_accuracy:
        _target_: xlm.metrics.MetricWrapper
        name: token_accuracy
        metric:
          _target_: xlm.metrics.TokenAccuracy
        prefix: test/prediction
        update_fn: xlm.metrics.seq2seq_token_accuracy_update_fn
    \end{minted}
\caption{Metric related entries in \bashline{configs/model_type/ilm.yaml} config file for the ILM Model.}
\label{fig:metrics}
\end{figure}

\subsection{Logging Predictions}
The model predictions for validation and test sets are logged under the \bashline{logs/runs} directory. The configuration for this is specified using the \bashline{log_predictions} key, and \xlm{}'s \bashline{xlm.log_predictions.LogPredictions} component should be used. The logger file contains a \textbf{text} field, which contains the prefix and generated sequence, and a \textbf{truth} field, which contains the ground-truth sequence. Predictions can be logged to a local file, trainer loggers, or the console by using the values 'file', 'logger', or 'console'.
\begin{figure}[H]
\begin{minted}{yaml}
log_predictions:
  _target_: xlm.log_predictions.LogPredictions
  fields_to_keep_in_output:
    - text
    - truth
  inject_target: target_ids
  writers:
    - file
    - logger
\end{minted}
\caption{The entries for logging predictions in the \bashline{experiment} config file.}
\label{fig:loggers}
\end{figure}

\section{FAQs}\label{sec:faqs}
\subsection{How to add a new task/dataset?}
\label{sec:appendix_new_datasets}
To add a new task, one must prepare the corresponding datasets in a Hugging Face (HF) dataset compatible format. Depending on the use case, separate dataset configuration (.yaml) files can be created for each stage (train/val/test/pred), providing the flexibility to process the same dataset differently or to use entirely different datasets across stages. The HF-downloadable dataset can be specified using the \bashline{full_name} key, and the \bashline{xlm.datamodule.DatasetManager} from \xlm{} can be reused for dataset manager instantiation. This manager automatically handles downloading, caching, preprocessing, and data-loading operations according to the dataset configuration entries. Alternatively, datasets can be used locally without being uploaded to the HF Hub by employing \bashline{xlm.datamodule.LocalDatasetManager}. Once these configuration files are prepared, they must be registered in the \bashline{configs/datamodule/MODEL.yaml} (Figure \ref{fig:config}).
\subsubsection{Implementing custom \datasetmanager{}}
In addition to XLM’s datasetmanager component, one can implement a custom datasetmanager for greater flexibility by inheriting from it. The necessary methods can be overridden—for example, the \bashline{_download} method shown below, where custom logic can be added to read and process the required type of file format.
\begin{minted} {python}
from xlm.datamodule import DatasetManager
import datasets

class CustomDatasetManager(DatasetManager):

    def _download(self) -> datasets.Dataset:
        ...
        return dataset
\end{minted}
The component can be configured along with mentioning the necessary arguments by adding \bashline{configs/datasets/custom_dataset_train.yaml}
\begin{minted}{yaml}
_target_: CustomDatasetManager
... # other arguments
\end{minted}
This config can then be registered in the \bashline{configs/datamodule/MODEL.yaml} file
\begin{minted}{yaml}
defaults:
-/datasets@datamodule.dataset_managers.train.lm: custom_dataset_train
... # other dataset and collator entries
\end{minted}

\section{Troubleshooting}\label{sec:troubleshooting}

\subsection{Hydra Errors}

\paragraph{\bashline{Unable to find a package ...} error by Hydra:}
See the name of the package in the error message. For example, if you encounter \bashline{Unable to find or instantiate abc.xyz.MyClass}, first try to import it manually in the Python interpreter: \bashline{python -c "from abc.xyz import MyClass"}.

\paragraph{Hydra Composition Errors:} First check the Hydra documentation \url{https://hydra.cc/docs/intro/}. If the error persists, write a single experiment config without using defaults list for components.

\input{drafts/eacl_demo/sections/05_aux}

%% file: drafts/eacl_demo/sections/05_aux.tex
\section{Additional Features}
\label{sec:auxiliary_features}
\subsection{Modules}\label{para:modules} The library, under the \textit{models} component, provides several architectural implementations that can be used to easily build diverse model backbones for prevalent non-autoregressive workflows in the literature. 
We provide modules for standard decoder only transformer, {Diffusion Transformers (\citealp{DiT}) , rotary embedder (\citealp{rope}), time embedder, adaptive layer normalization layers (\citealp{DiT}}, and some standard {noise schedulers}. 
They are available under \pythonline{xlm.modules}. 

\subsection{Push to hub} \label{sec:push_to_hub}
The library provides a \pythonline{push_to_hub} command that uploads trained models to the Hugging Face Hub by reconstructing the complete training environment (datamodule, tokenizer and model architecture) from Hydra configurations.
\begin{minted}{bash}
$ xlm-push-to-hub experiment=[CONFIG_PATH] +hub_checkpoint_path=[CKPT_PATH] +hub.repo_id=[HUB_PATH]
\end{minted}
The environment variable \bashline{HF_HUB_KEY} must be assigned a valid Hugging Face access token.
\subsection{Callbacks} The library extends the PyTorch Lightning callback infrastructure, enabling modular components to integrate with the training cycle (e.g., per-batch updates, validation hooks) in a decoupled manner. It provides a set of extensible callbacks, such as an Exponential Moving Average callback \pythonline{EMACallback} for maintaining smoothed evaluation weights, a Checkpoint Monitoring callback \pythonline{ModelCheckpoint} and a Performance Monitoring callback \pythonline{SpeedMonitorCallback} for tracking training speed and identifying bottlenecks. The callback config file names (\xlm{}'s or custom callbacks) can be mentioned in the following way to override the default callbacks:

\begin{minted}{yaml}
defaults:
  - override /callbacks:
    - ema
    - speed_monitor
    - checkpoint_monitor
\end{minted}
\subsection{Checkpointing} The library provides a checkpointing system that saves training state (model weights, optimizer state, and training progress) to enable recovery from failures and long-running training jobs. It integrates with PyTorch Lightning’s built-in \pythonline{ModelCheckpoint} to save the best-performing model. It also supports frequent lightweight checkpoints using a \pythonline{ThinningCheckpoint} callback that retains only milestone intervals to save storage and an \pythonline{OnExceptionCheckpoint} callback that preserves state during crashes.

\input{drafts/eacl_demo/sections/06_tasks}

\section{Planned Features}\label{app:planned features}
\paragraph{Non-text datasets}
Non-autoregressive sequence generation is useful for non-text tasks like molecule generation~\citep{irwin2005zinc,ruddigkeit2012enumeration}, path planning, etc. We plan on adding support for external non-text datasets in the future

\paragraph{New Models}
We plan to add support for newer models~\citep{editflow,flexmdm}.

\paragraph{FlexAttention}
Pytorch 2.5 introduces FlexAttention~\citep{flexattention},  dynamically compiled attention layer which allows fast attention with arbitrary masks. This can be very useful for non-autoregressive sequence generation as it can allow sequence packing eliminating the need for padding even for non-autoregressive models.

\section{Resources}\label{app:resources}
Table \ref{tab:resources} lists the resources provided through this paper.

\begin{table*}
    \caption{Resources}
    \label{tab:resources}
    \small
    \begin{tabular}{p{1.8cm}p{1.2cm}p{2.5cm}p{4cm}p{2cm}}
        \toprule
        Name & Type & Description & Link & License\\
        \midrule
        \texttt{StarEasy} & Dataset & Synthetic star graph dataset & \url{https://github.com/facebookresearch/xlm/tree/main/xlm/datasets/star_easy} & CC-BY-NC-SA\\
        \texttt{StarMedium} & Dataset & Synthetic star graph dataset & \url{https://github.com/facebookresearch/xlm/tree/main/xlm/datasets/star_medium} & CC-BY-NC-SA\\
        \texttt{StarHard} & Dataset & Synthetic star graph dataset & \url{https://github.com/facebookresearch/xlm/tree/main/xlm/datasets/star_hard} & CC-BY-NC-SA\\
        \xlmcore{} & Python Package & \xlmcore{} package for non-autoregressive language modeling & \url{https://pypi.org/project/xlm-core/} & MIT\\
        \xlmmodels{} & Python Package & Companion package for \xlmcore{} containing model implementations & \url{https://pypi.org/project/xlm-models/} & MIT\\
        \bottomrule
    \end{tabular}
\end{table*}

%% file: drafts/eacl_demo/sections/06_tasks.tex
\section{Preconfigured Tasks and Models}
\label{sec:preconfigured-tasks}
\paragraph{Star Graphs}
The library provides three synthetic datasets that involve generating the path from a designated start node to a target node on star-shaped graphs~\citep{pmlr-v235-bachmann24a, ilm}. 
The three variants follow the naming convention and construction of \citet{ilm}.
\texttt{StarEasy} contains symmetric graphs with the start node fixed at the junction, while \texttt{StarMedium} \& \texttt{StarHard} introduce asymmetric structures with variable arm lengths and start nodes that may lie off the junction.

\paragraph{Language Modeling}
For text generation, we provide training and testing config setup for two datasets - LM1B, a large-scale corpus from the news domain, consisting of short text sequences (typically 2–3 sentences), and OpenWebText, which are widely used to benchmark the performance of small language models.

\paragraph{Models} We benchmark and release three preconfigured models: ARLM, MDLM and ILM. \footnote{We are working on benchmarking newer models, which will be released soon.}

%% file: main.bbl
\begin{thebibliography}{24}
\providecommand{\natexlab}[1]{#1}

\bibitem[{Bachmann and Nagarajan(2024)}]{pmlr-v235-bachmann24a}
Gregor Bachmann and Vaishnavh Nagarajan. 2024.
\newblock \href {https://proceedings.mlr.press/v235/bachmann24a.html} {The pitfalls of next-token prediction}.
\newblock In \emph{Proceedings of the 41st International Conference on Machine Learning}, volume 235 of \emph{Proceedings of Machine Learning Research}, pages 2296--2318. PMLR.

\bibitem[{Chelba et~al.(2014)Chelba, Mikolov, Schuster, Ge, Brants, Koehn, and Robinson}]{chelba2014billionwordbenchmarkmeasuring}
Ciprian Chelba, Tomas Mikolov, Mike Schuster, Qi~Ge, Thorsten Brants, Phillipp Koehn, and Tony Robinson. 2014.
\newblock \href {https://arxiv.org/abs/1312.3005} {One billion word benchmark for measuring progress in statistical language modeling}.
\newblock \emph{Preprint}, arXiv:1312.3005.

\bibitem[{Dong et~al.(2024)Dong, Feng, Guessous, Liang, and He}]{flexattention}
Juechu Dong, Boyuan Feng, Driss Guessous, Yanbo Liang, and Horace He. 2024.
\newblock \href {https://arxiv.org/abs/2412.05496} {Flex attention: A programming model for generating optimized attention kernels}.
\newblock \emph{Preprint}, arXiv:2412.05496.

\bibitem[{Falcon and {The PyTorch Lightning team}(2019)}]{pytorch_lightning}
William Falcon and {The PyTorch Lightning team}. 2019.
\newblock \href {https://doi.org/10.5281/zenodo.3828935} {{PyTorch Lightning}}.

\bibitem[{Gamma et~al.(1995)Gamma, Helm, Johnson, and Vlissides}]{design_patterns}
Erich Gamma, Richard Helm, Ralph Johnson, and John Vlissides. 1995.
\newblock \emph{Design patterns: elements of reusable object-oriented software}.
\newblock Addison-Wesley Longman Publishing Co., Inc., USA.

\bibitem[{Gardner et~al.(2017)Gardner, Grus, Neumann, Tafjord, Dasigi, Liu, Peters, Schmitz, and Zettlemoyer}]{Gardner2017AllenNLP}
Matt Gardner, Joel Grus, Mark Neumann, Oyvind Tafjord, Pradeep Dasigi, Nelson~F. Liu, Matthew Peters, Michael Schmitz, and Luke~S. Zettlemoyer. 2017.
\newblock \href {https://arxiv.org/abs/arXiv:1803.07640} {Allennlp: A deep semantic natural language processing platform}.

\bibitem[{Gulrajani and Hashimoto(2023)}]{plaid}
Ishaan Gulrajani and Tatsunori Hashimoto. 2023.
\newblock \href {https://openreview.net/forum?id=e2MCL6hObn} {Likelihood-based diffusion language models}.
\newblock In \emph{Thirty-seventh Conference on Neural Information Processing Systems}.

\bibitem[{Havasi et~al.(2025)Havasi, Karrer, Gat, and Chen}]{editflow}
Marton Havasi, Brian Karrer, Itai Gat, and Ricky T.~Q. Chen. 2025.
\newblock \href {https://doi.org/10.48550/arXiv.2506.09018} {Edit {{Flows}}: {{Flow Matching}} with {{Edit Operations}}}.
\newblock \emph{Preprint}, arXiv:2506.09018.

\bibitem[{Irwin and Shoichet(2005)}]{irwin2005zinc}
John~J Irwin and Brian~K Shoichet. 2005.
\newblock Zinc- a free database of commercially available compounds for virtual screening.
\newblock \emph{Journal of chemical information and modeling}, 45(1):177--182.

\bibitem[{Kim et~al.(2025)Kim, {Cheuk-Kit}, {Domingo-Enrich}, Du, Kakade, Ngotiaoco, Chen, and Albergo}]{flexmdm}
Jaeyeon Kim, Lee {Cheuk-Kit}, Carles {Domingo-Enrich}, Yilun Du, Sham Kakade, Timothy Ngotiaoco, Sitan Chen, and Michael Albergo. 2025.
\newblock \href {https://doi.org/10.48550/arXiv.2509.01025} {Any-{{Order Flexible Length Masked Diffusion}}}.
\newblock \emph{Preprint}, arXiv:2509.01025.

\bibitem[{Kwon et~al.(2023)Kwon, Li, Zhuang, Sheng, Zheng, Yu, Gonzalez, Zhang, and Stoica}]{kwon2023efficient}
Woosuk Kwon, Zhuohan Li, Siyuan Zhuang, Ying Sheng, Lianmin Zheng, Cody~Hao Yu, Joseph~E. Gonzalez, Hao Zhang, and Ion Stoica. 2023.
\newblock Efficient memory management for large language model serving with pagedattention.
\newblock In \emph{Proceedings of the ACM SIGOPS 29th Symposium on Operating Systems Principles}.

\bibitem[{Li et~al.(2025)Li, Zheng, and Jiang}]{llm_api_documentation_accuracy}
Pengfei Li, Qichang Zheng, and Ziyi Jiang. 2025.
\newblock \href {https://doi.org/10.63575/} {An empirical study on the accuracy of large language models in api documentation understanding: A cross-programming language analysis}.
\newblock \emph{Journal of Computing Innovations and Applications}, 3(2):1–14.

\bibitem[{Lightning-AI(2023)}]{litgpt-2023}
Lightning-AI. 2023.
\newblock Litgpt.
\newblock \url{https://github.com/Lightning-AI/litgpt}.

\bibitem[{OLMo et~al.(2024)OLMo, Walsh, Soldaini, Groeneveld, Lo, Arora, Bhagia, Gu, Huang, Jordan, Lambert, Schwenk, Tafjord, Anderson, Atkinson, Brahman, Clark, Dasigi, Dziri, Guerquin, Ivison, Koh, Liu, Malik, Merrill, Miranda, Morrison, Murray, Nam, Pyatkin, Rangapur, Schmitz, Skjonsberg, Wadden, Wilhelm, Wilson, Zettlemoyer, Farhadi, Smith, and Hajishirzi}]{olmo20242olmo2furious}
Team OLMo, Pete Walsh, Luca Soldaini, Dirk Groeneveld, Kyle Lo, Shane Arora, Akshita Bhagia, Yuling Gu, Shengyi Huang, Matt Jordan, Nathan Lambert, Dustin Schwenk, Oyvind Tafjord, Taira Anderson, David Atkinson, Faeze Brahman, Christopher Clark, Pradeep Dasigi, Nouha Dziri, and 21 others. 2024.
\newblock \href {https://arxiv.org/abs/2501.00656} {2 olmo 2 furious}.
\newblock \emph{Preprint}, arXiv:2501.00656.

\bibitem[{Ott et~al.(2019)Ott, Edunov, Baevski, Fan, Gross, Ng, Grangier, and Auli}]{ott-etal-2019-fairseq}
Myle Ott, Sergey Edunov, Alexei Baevski, Angela Fan, Sam Gross, Nathan Ng, David Grangier, and Michael Auli. 2019.
\newblock \href {https://doi.org/10.18653/v1/N19-4009} {fairseq: A fast, extensible toolkit for sequence modeling}.
\newblock In \emph{Proceedings of the 2019 Conference of the North {A}merican Chapter of the Association for Computational Linguistics (Demonstrations)}, pages 48--53, Minneapolis, Minnesota. Association for Computational Linguistics.

\bibitem[{Paszke et~al.(2019)Paszke, Gross, Massa, Lerer, Bradbury, Chanan, Killeen, Lin, Gimelshein, Antiga, Desmaison, Köpf, Yang, DeVito, Raison, Tejani, Chilamkurthy, Steiner, Fang, Bai, and Chintala}]{pytorch}
Adam Paszke, Sam Gross, Francisco Massa, Adam Lerer, James Bradbury, Gregory Chanan, Trevor Killeen, Zeming Lin, Natalia Gimelshein, Luca Antiga, Alban Desmaison, Andreas Köpf, Edward Yang, Zach DeVito, Martin Raison, Alykhan Tejani, Sasank Chilamkurthy, Benoit Steiner, Lu~Fang, and 2 others. 2019.
\newblock \href {https://arxiv.org/abs/1912.01703} {Pytorch: An imperative style, high-performance deep learning library}.
\newblock \emph{Preprint}, arXiv:1912.01703.

\bibitem[{Patel et~al.(2025)Patel, Sahoo, Amballa, Naseem, Rudner, and McCallum}]{ilm}
Dhruvesh Patel, Aishwarya Sahoo, Avinash Amballa, Tahira Naseem, Tim~GJ Rudner, and Andrew McCallum. 2025.
\newblock Insertion language models: Sequence generation with arbitrary-position insertions.
\newblock \emph{arXiv preprint arXiv:2505.05755}.

\bibitem[{Peebles and Xie(2023)}]{DiT}
William Peebles and Saining Xie. 2023.
\newblock Scalable diffusion models with transformers.
\newblock In \emph{Proceedings of the IEEE/CVF international conference on computer vision}, pages 4195--4205.

\bibitem[{Ruddigkeit et~al.(2012)Ruddigkeit, Van~Deursen, Blum, and Reymond}]{ruddigkeit2012enumeration}
Lars Ruddigkeit, Ruud Van~Deursen, Lorenz~C Blum, and Jean-Louis Reymond. 2012.
\newblock Enumeration of 166 billion organic small molecules in the chemical universe database gdb-17.
\newblock \emph{Journal of chemical information and modeling}, 52(11):2864--2875.

\bibitem[{Sahoo et~al.(2024)Sahoo, Arriola, Gokaslan, Marroquin, Rush, Schiff, Chiu, and Kuleshov}]{sahooSimpleEffectiveMasked2024}
Subham~Sekhar Sahoo, Marianne Arriola, Aaron Gokaslan, Edgar~Mariano Marroquin, Alexander~M. Rush, Yair Schiff, Justin~T. Chiu, and Volodymyr Kuleshov. 2024.
\newblock \href {https://openreview.net/forum?id=L4uaAR4ArM} {Simple and {{Effective Masked Diffusion Language Models}}}.
\newblock In \emph{The {{Thirty-eighth Annual Conference}} on {{Neural Information Processing Systems}}}.

\bibitem[{Su et~al.(2024)Su, Ahmed, Lu, Pan, Bo, and Liu}]{rope}
Jianlin Su, Murtadha Ahmed, Yu~Lu, Shengfeng Pan, Wen Bo, and Yunfeng Liu. 2024.
\newblock Roformer: Enhanced transformer with rotary position embedding.
\newblock \emph{Neurocomputing}, 568:127063.

\bibitem[{von Werra et~al.(2020)von Werra, Belkada, Tunstall, Beeching, Thrush, Lambert, Huang, Rasul, and Gallouédec}]{vonwerra2022trl}
Leandro von Werra, Younes Belkada, Lewis Tunstall, Edward Beeching, Tristan Thrush, Nathan Lambert, Shengyi Huang, Kashif Rasul, and Quentin Gallouédec. 2020.
\newblock Trl: Transformer reinforcement learning.
\newblock \url{https://github.com/huggingface/trl}.

\bibitem[{Wolf et~al.(2020)Wolf, Debut, Sanh, Chaumond, Delangue, Moi, Cistac, Rault, Louf, Funtowicz, Davison, Shleifer, von Platen, Ma, Jernite, Plu, Xu, Scao, Gugger, Drame, Lhoest, and Rush}]{wolf2020huggingfacestransformersstateoftheartnatural}
Thomas Wolf, Lysandre Debut, Victor Sanh, Julien Chaumond, Clement Delangue, Anthony Moi, Pierric Cistac, Tim Rault, Rémi Louf, Morgan Funtowicz, Joe Davison, Sam Shleifer, Patrick von Platen, Clara Ma, Yacine Jernite, Julien Plu, Canwen Xu, Teven~Le Scao, Sylvain Gugger, and 3 others. 2020.
\newblock \href {https://arxiv.org/abs/1910.03771} {Huggingface's transformers: State-of-the-art natural language processing}.
\newblock \emph{Preprint}, arXiv:1910.03771.

\bibitem[{Yadan(2019)}]{Yadan2019Hydra}
Omry Yadan. 2019.
\newblock \href {https://github.com/facebookresearch/hydra} {Hydra - a framework for elegantly configuring complex applications}.
\newblock Github.

\end{thebibliography}
